\documentclass[10pt,twocolumn,letterpaper]{article}

\usepackage[pagenumbers]{wacv} 

\usepackage{graphicx}
\usepackage{amsmath}
\usepackage{amssymb}
\usepackage{booktabs}

\usepackage[pagebackref,breaklinks,colorlinks]{hyperref}

\usepackage[capitalize]{cleveref}
\crefname{section}{Sec.}{Secs.}
\Crefname{section}{Section}{Sections}
\Crefname{table}{Table}{Tables}
\crefname{table}{Tab.}{Tabs.}

\def\wacvPaperID{1272} 
\def\confName{WACV}
\def\confYear{2024}

\begin{document}

\title{LInKs ``Lifting Independent Keypoints" - Partial Pose Lifting for Occlusion Handling with Improved Accuracy in 2D-3D Human Pose Estimation}

\author{Peter Hardy \& Hansung Kim\\
University of Southampton\\
Vision Learning and Control, ECS\\
{\tt\small p.t.d.hardy@soton.ac.uk}
}
\maketitle

\begin{abstract}
We present LInKs, a novel unsupervised learning method to recover 3D human poses from 2D kinematic skeletons obtained from a single image, even when occlusions are present. Our approach follows a unique two-step process, which involves first lifting the occluded 2D pose to the 3D domain, followed by filling in the occluded parts using the partially reconstructed 3D coordinates. This lift-then-fill approach leads to significantly more accurate results compared to models that complete the pose in 2D space alone. Additionally, we improve the stability and likelihood estimation of normalising flows through a custom sampling function replacing PCA dimensionality reduction previously used in prior work. Furthermore, we are the first to investigate if different parts of the 2D kinematic skeleton can be lifted independently which we find by itself reduces the error of current lifting approaches. We attribute this to the reduction of long-range keypoint correlations. In our detailed evaluation, we quantify the error under various realistic occlusion scenarios, showcasing the versatility and applicability of our model. Our results consistently demonstrate the superiority of handling all types of occlusions in 3D space when compared to others that complete the pose in 2D space. Our approach also exhibits consistent accuracy in scenarios without occlusion, as evidenced by a 7.9\% reduction in reconstruction error compared to prior works on the Human3.6M dataset. Furthermore, our method excels in accurately retrieving complete 3D poses even in the presence of occlusions, making it highly applicable in situations where complete 2D pose information is unavailable.
\end{abstract}

\section{Introduction}
\label{sec:intro}
Human pose estimation (HPE) is an important task in computer vision with applications in various fields, such as human-computer interaction, augmented reality, and healthcare \cite{human_robot, KUMARAPU202116, healthcare}. However, recovering 3D human poses from a single image is known to be an ill-posed inverse problem, as multiple different 2D poses can correspond to the same 3D pose. Traditional approaches in 2D-3D HPE have therefore required either multiple views of the subject or a depth sensor, which limits their applicability in real-world scenarios where multiple views or depth sensors may be unobtainable. In recent years, unsupervised learning methods have shown promising results in 3D HPE from single images, where the model learns to extract 3D pose information from 2D poses without any 3D annotations \cite{kudo2018unsupervised, Yu_2021_ICCV, amazon_paper_2, Wandt_2022_CVPR, Drover_2018_ECCV_Amazon}. However, these approaches operate by lifting the entire 2D kinematic skeleton during training which has several limitations. Firstly, they do not account for occlusions or 2D keypoint detection errors, making them unable to handle incomplete or adjust to bad, information. As a result, the omission of just a single joint will lead to these models being unable to work. Secondly, lifting the entire 2D kinematic skeleton may result in long-range correlations between anatomically unassociated keypoints being learned during training. As a result, the 2D coordinate of one keypoint may inappropriately influence the 3D estimate of another that is far away. This can lead to inaccurate pose estimations, especially in relation to complex human poses with multiple degrees of freedom. Thus, there is a need for more robust methods that can effectively handle occlusions and account for the complex dependencies between keypoints in the estimation process. Our paper, therefore, makes the following important contributions:
\begin{figure*}[t]
\centering
\includegraphics[width=\textwidth]{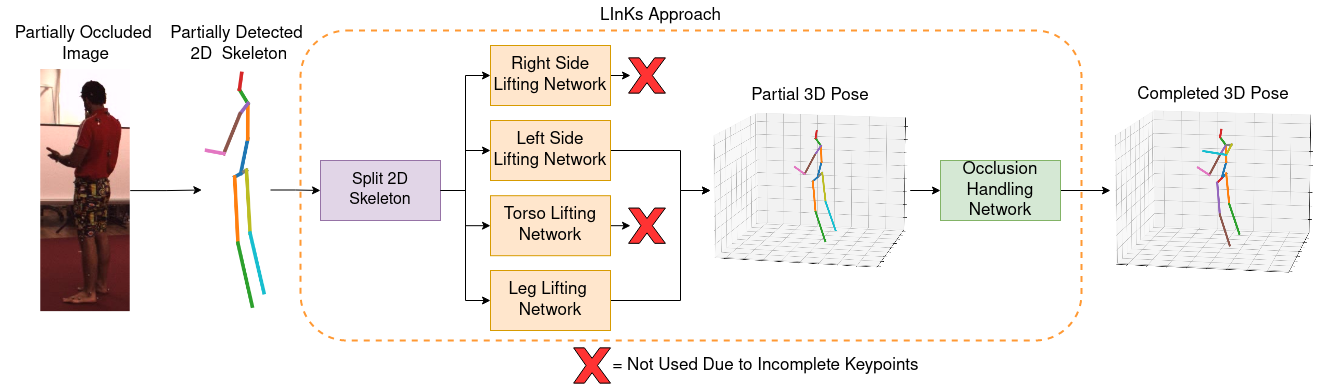} 
\caption{Overview of the lifting and occlusion handling of the LInKs approach. A partially detected 2D skeleton is obtained from an occluded image. The 2D skeleton is broken into its corresponding torso, legs, and left and right-hand side keypoints which are provided to their respective lifting networks whose outputs are combined to obtain a partial 3D pose. This partial 3D pose is provided to an occlusion handling network which predicts the missing keypoints giving us our full 3D pose. In the above scenario the right arm is occluded therefore the right-hand side and torso lifting networks are not used due to incomplete 2D keypoint information.}
\label{end_to_end_flow}
\end{figure*}
\subsection{Lift then Fill Two-Stage Approach} Prior 2D-3D lifting research \cite{Wandt_2022_CVPR, Wandt2019RepNet, amazon_paper_2, Drover_2018_ECCV_Amazon, Yu_2021_ICCV, Martinez_2017_ICCV} have mainly assumed that 2D pose detection models can accurately capture complete 2D poses from single images. We argue that this assumption may be flawed. By using OpenPose on videos from the Human3.6M dataset, we found that complete poses were detected in only 54.8\% and 35.5\% of frames from the front and rear cameras respectively, with an average full-pose retrieval rate of 45.1\%. Put plainly, the previously mentioned approaches would only function in 45.1\% of all available frames. This can be attributed to self-occlusion or 2D detection error, which results in missing limbs or keypoints in various scenes. To address this, for the first stage of our approach, we lifted different segments of the 2D human pose independently of one another to obtain a partial 3D pose in cases of occlusion. In the next stage, this partial 3D pose was then seen by an occlusion handling network which filled in the missing coordinates to retrieve a complete 3D pose. An overview of these stages can be seen in Figure \ref{end_to_end_flow}. Our lift-then-fill approach has multiple benefits over handling the occlusion at the 2D stage namely:
\begin{itemize}
    \item \textbf{Consistency with Human Anatomy:} Human joints have specific ranges of motion and dependencies on neighbouring joints, which are more accurately captured in 3D. By choosing to fill in the occluded joints solely in 2D space we may obtain poses that violate anatomical constraints and look unnatural once lifted into 3D.
    \item \textbf{Reducing Model Complexity:} As mentioned earlier, a single 3D pose can have various corresponding 2D representations. Consequently, a 2D occlusion model must learn the multiple potential 2D pose interpretations stemming from a single 3D pose. However, addressing this challenge becomes more manageable when working with partial 3D poses due to the inherent consistency of 3D poses during rotations.
    \item \textbf{Reducing Error Propagation:} By tackling occlusion in 3D space, we remove any errors or inaccuracies that would have otherwise been propagated to the lifting network. Even if a 2D occlusion handling network managed to infer the 2D location of the occluded joint relatively well, the lifting network may still struggle to accurately convert them to the correct 3D coordinates. This is especially true if the 2D occlusion handling network introduces subtle errors or noise during its process.
\end{itemize}
Further results of our OpenPose analysis investigating the percentage of full poses detected in the Human3.6M dataset can be found in the supplementary material.
\subsection{Normalising Flow Sampling} Since unsupervised learning approaches can benefit more from increased data than their supervised counterparts, it is beneficial to exploit multiple data sources \cite{Drover_2018_ECCV_Amazon}. To address the challenges of limited data sources within 3DHPE, we leverage the power of normalising flows both in terms of their generative and likelihood capabilities. Put simply, normalising flows are invertible functions that learn to map between simple and complicated distributions. By sampling from the estimated distribution during training, we can therefore generate additional data, allowing us to effectively exploit the limited data we have available. However, we found that random sampling led to the formation of impossible and unnatural 2D poses. To solve this, we introduced a new sampling approach that allowed the flow to generate and learn more meaningful representations of the data.

\section{Related Work}
\label{related_work}
There currently exist two main avenues of deep-learning research for 3D HPE. The first learns the mapping of 3D joints directly from a 2D image \cite{Pavlakos2017CoarsetoFineVP, 8795015, mono-3dhp2017, Li2015MaximumMarginSL, TomeD2017LftD}. The second builds upon an accurate intermediate 2D pose estimate, with the 2D pose obtained from an image using techniques such as Stacked-Hourglass Architectures \cite{stacked_hourglass} or Part Affinity Fields \cite{open_pose} and lifting this 2D pose to 3D. This paper focuses on the latter 2D to 3D lifting avenue which can be organized into the following categories:
\subsection{Full and Weak Supervision} 
Fully supervised approaches seek to learn mappings from paired 2D-3D data, which contain ground truth 2D locations of keypoints and their corresponding 3D coordinates. In comparison, weakly-supervised approaches do not use corresponding 2D-3D data, instead using either augmented 3D data during training or unpaired 2D-3D data. Martinez \etal \cite{Martinez_2017_ICCV} introduced the first baseline fully connected regression model that learned 3D coordinates from their relative 2D locations in images. This fully supervised work also introduced the residual block architecture which has been adopted as the standard in the field of 2D-3D pose estimation. Yang \etal \cite{Yang20183DHP} used an adversarial approach, with a critic network that compared their lifted ``in-the-wild" 3D poses against ground-truth 3D poses obtained from a controlled setting. Wandt and Rosenhahn \cite{Wandt2019RepNet} also followed this line of research but transformed their predicted and ground truth 3D poses into a kinematic chain \cite{Wandt2018AKC} before being seen by a Wasserstein critic network \cite{improved_wasserstein} which reduced the overfitting present in direct 3D correspondence. In contrast, Drover \etal \cite{Drover_2018_ECCV_Amazon} investigated if 3D poses can be learned through 2D self-consistency alone, rotating a predicted 3D pose and reprojecting it back into 2D before passing it through the model for comparison. This led to the discovery that an adversarial 2D critic network was needed as self-consistency alone was not sufficient. Conversely, Mehta \etal \cite{mono-3dhp2017} and Kundu \etal \cite{kundu2020self} took a transfer learning approach to 3D HPE. They used mixed 2D-3D labels and images in a unified network to predict the 3D pose in scenarios where no 3D data was available. Compared to fully-supervised approaches, weakly supervised methods generalise better to unseen pose scenarios. However, they still struggle with poses that are very different from those within the training data. 
\subsection{Unsupervised} 
Unsupervised approaches do not utilise any 3D data during training, unpaired or otherwise. Kudo \etal \cite{kudo2018unsupervised} introduced one of the first unsupervised adversarial networks, utilising random reprojections and a 2D critic network under the assumption that any predicted 3D pose, once rotated and reprojected, would still produce a believable 2D pose. Chen \etal \cite{amazon_paper_2} expanded this work and introduced an unsupervised adversarial approach with the rotational consistency cycle presented by Drover \etal \cite{Drover_2018_ECCV_Amazon}. Yu \etal \cite{Yu_2021_ICCV} built upon this work further. They highlighted that the temporal constraints introduced in Chen \etal \cite{amazon_paper_2} may hinder a model's performance due to balancing multiple training objectives simultaneously, and proposed splitting the problem into both a lifting and temporal scale estimation module. More recently, Wandt \etal \cite{Wandt_2022_CVPR} dispensed with the critic network and relied instead on the pose likelihood of a pre-trained normalising flow, allowing for a more interpretable loss during training. However, to stabilise the normalising flow during training they performed PCA on the distribution of 2D poses, reducing the dimensionality as well as the information present. Importantly, PCA may have hindered one of the primary advantages of normalising flows, their ability to learn exact bijective mappings between two different data distributions. Therefore, in our approach, we removed dimensionality reduction and replaced it with generative sampling from the latent distribution during training. Additionally, all previously mentioned approaches lift the entire 2D pose to 3D during training making them insufficient in occlusion scenarios. To our knowledge, our study is the first to utilise completely unsupervised networks to explore the feasibility of lifting different sections of a 2D pose to 3D independently for the purpose of obtaining partial 3D poses in occlusion scenarios.
\subsection{Unsupervised 2D-3D Occlusion Handling}
In our systemically undertaken review of published literature, only one prior work was identified that investigated unsupervised 2D-3D lifting from a single image, OCR-Pose by Wang \etal \cite{OCR_POSE}. OCR-Pose incorporates two modules: a topology invariant contrastive learning (TiCLR) module and a view equivariant contrastive learning (VeCLR) module. The TiCLR module aims to bridge the gap between an occluded and unoccluded 2D pose while the VeCLR module encourages the lifting network to produce consistent 3D skeletons from multiple viewpoints.  However, by completing the occluded pose solely in 2D space the the accuracy of their approach is limited as previously discussed. We suggest a different strategy. We attempted to generate a partial 3D pose using the available 2D keypoints and subsequently fill in the occluded parts in 3D space. We find our alternative methodology reduced the error and we also simulate more realistic occlusion scenarios than OCR-Pose during our evaluation.
\section{Method}
In this section, we present our unsupervised learning approach for lifting 2D poses to 3D. Our 2D poses consisted of $N$ keypoints, $(x_i, y_i)$, $i = 1...N$, where the root keypoint, located at the origin $(0, 0)$, was the midpoint between the left and right hip joint. As monocular images are subject to fundamental perspective ambiguity, deriving absolute depth from a single view is not possible \cite{depth, amazon_paper_2}. To address this, we adopted the practice of fixing the distance of the pose from the camera at a constant $c$ units and normalising such that the average distance from the head to the root keypoint was $\frac{1}{c}$ units in 2D, with $c$ being set to 10 as is consistent with previous research \cite{amazon_paper_2, Wandt_2022_CVPR, Yu_2021_ICCV}.
\subsection{Independent Lifting Networks}
Our lifting networks, inspired by \cite{Martinez_2017_ICCV}, were trained to predict the 3D depth off-set ($\hat{d}$) from the poses root keypoint for each 2D keypoint $(x, y)$. The final 3D location of a specific keypoint, $\mathbf{x}_i$, was then obtained via perspective projection:
\begin{equation}
\begin{split}
\mathbf{x}_i &= (x_i\hat{z}_i, y_i\hat{z}_i, \hat{z}_i), \\
\mathbf{where } \quad \hat{z}_i &= \max(1, \hat{d}_i + c).
\end{split}
\end{equation}
where $d_i$ was our models' depth-offset prediction for keypoint $i$. We used independent lifting networks for the legs, torso and left and right side keypoints and adopted this arrangement as during our analysis with OpenPose, we noticed that single-limb occlusions or same-side upper and lower limb occlusions were the most common.
\subsection{Normalising Flows with Generative Sampling}
\begin{figure}[h]
\centering
\includegraphics[width=0.99\columnwidth]{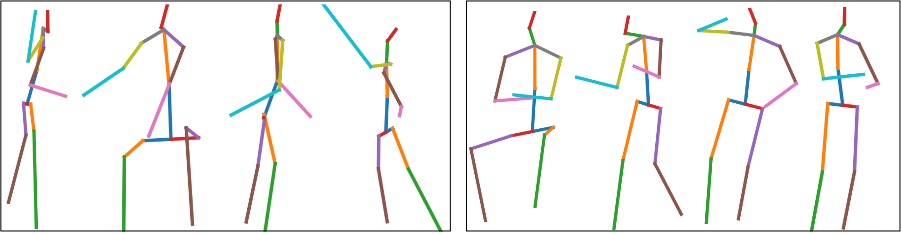}
\caption{Showing 2D poses drawn from the learned latent space of our normalising flows via random sampling (left) and our improved sampling (right). Note how random sampling leads to abnormal poses such as limbs being too long and unnaturally bent}
\label{samples_drawn}
\end{figure}

Normalising flows are widely used generative models for estimating complex probability distributions, including those of 2D poses. In our scenario, $x$ is our 2D pose data and $z$ is the latent variable (Gaussian in our experiment) such that $x = f(z)$, where $f$ is the flow function, the probability density function of the input data can be computed by using the change of variables formula:
\begin{equation}
p_{X}(x) = p_{Z}(z) \left| \det \left( \frac{\partial f(z)}{\partial z} \right) \right|^{-1}
\end{equation}
where $p_{Z}(z)$ is the probability density function of the latent variable $z$ and $p_{X}(x)$ is the probability density function of the input data $x$ (the 2D poses). The determinant of the Jacobian matrix of the flow function, denoted by $\det \left( \frac{\partial f(z)}{\partial z} \right)$, serves as a normalisation term that accounts for the change in volume induced by the transformation $f$. Previously PCA was used to reduce the dimensionality of 2D poses prior to training the normalising flow \cite{Wandt_2022_CVPR}. The reason for this was the high dimensionality of input data which prevented flow optimisation. Using our approach, we found that this step was not needed as we could incorporate pose sampling during the training of our flows to increase stability. Specifically, while training the flow to maximise the likelihood of the training data, we also maximised the likelihood of sampled 2D poses drawn from the estimated distribution. However, our analysis revealed that simple random sampling can cause the flow to collapse due to impossible 2D poses being generated. We posit that this is due to the interconnected nature of 2D keypoints which the flow may have trouble learning. To address this issue, we introduced a modified sampling strategy that first obtained the estimated distribution samples of the true 2D poses. We then augmented this sample by adding Gaussian noise, scaled by a factor of $\sigma$, and the Gaussian features of the current sample, before passing it through the inverse of the normalising flow to generate new 2D poses. This modification can be expressed as follows:
\begin{equation}
\label{sampling}
\textbf{x}_i' = f_{\theta}(\mathbf{z}_i + \sigma\mathbf{z}_i\mathcal{N}(0,1))
\end{equation}
where $\textbf{x}_i'$ is the sampled 2D pose, $f_{\theta}$ is the normalising flow with parameters $\theta$, $\textbf{z}_i$ is the estimated distribution sample of a true 2D pose, $\sigma$ is a scaling factor and $\mathcal{N}(0,1)$ is our random Gaussian noise. Examples of poses drawn from the latent distribution of our flow via random sampling and our own sampling approach can be seen in Figure \ref{samples_drawn}.
During training, we updated the parameters of our normalising flow to minimise the negative log-likelihood of both the sampled and ground-truth 2D poses, defined as:
\begin{equation}
\label{likelihood}
\mathcal{L}_{\theta} = - \frac{1}{N}\sum_{i=1}^N \left[ \log p_{\theta}(\textbf{x}_i') + \log p_{\theta}(\textbf{x}_i) \right]
\end{equation}
where $p_{\theta}$ is the probability density function estimated by the normalising flow, and $\textbf{x}_i$ is the ground-truth 2D pose. We trained a total of five normalising flows for our model. Four flows were trained on the leg, torso, and left and right side keypoints and were used for likelihood estimation during the training of each corresponding lifting network. The final normalising flow was trained on the entire pose and was specifically used for the generative sampling of new poses during the training of each lifting network and their respective flows.
\subsection{Independent Lifting, Rotational consistency and Re-projection Likelihood.}
Motivated by multi-view camera setups, where depth can be inferred through re-projection to another view, unsupervised learning utilises a virtual second view during training to mimic this property \cite{amazon_paper_2, Yu_2021_ICCV, Wandt_2022_CVPR, Drover_2018_ECCV_Amazon}. Given a 2D input pose $\mathbf{Y}_{2D}\in\mathbb{R}^{N \times 2}$, a corresponding 3D pose $\hat{\mathbf{Y}}_{3D}$ is obtained via perspective projection. $\hat{\mathbf{Y}}_{3D}$ is then rotated by a rotation matrix $\mathbf{R}$ to produce a new 3D pose. The azimuth angle of $\mathbf{R}$ is obtained by random sampling from a uniform distribution between $[-\pi, \pi]$, with the elevation angle learned during training as detailed in work by \cite{Wandt_2022_CVPR}. Subsequently, the rotated 3D pose is re-projected back to 2D using the projection matrix $\mathbf{P}$. This yields a new synthetic viewpoint of the pose, represented by the matrix $\tilde{\mathbf{Y}}_{2D}\in\mathbb{R}^{N \times 2}$. The new 2D pose is then split into the leg, torso, and left and right side keypoints before being given to our pre-trained normalising flows for density estimation. As a result, each normalising flow provides a likelihood value representing the probability of each 2D segment occurring within the learned distribution from the training dataset $(\mathcal{L}_{NF})$. To promote self-consistency during training, we passed $\tilde{\mathbf{Y}}_{2D}$ through the same lifting networks and obtain a new 3D pose from this virtual viewpoint  $\tilde{\mathbf{Y}}_{3D}$. We performed the inverse rotation $\mathbf{R}^{-1}$ on $\tilde{\mathbf{Y}}_{3D}$, and reprojected it back into 2D. Adopting this process enabled the original matrix of 2D keypoints $\mathbf{Y}_{2D}$ to be recreated, thereby facilitating our model to learn consistency. Specifically, our lifting networks sought to minimize the following component:
\begin{equation}
\label{twod_loss}
\mathcal{L}_{2D}=\left|\mathbf{Y}_{2D}-\mathbf{P}\mathbf{R}^{-1}\tilde{\mathbf{Y}}_{3D}\right|
\end{equation}
As we lifted different 2D keypoints independently, each lifter received its own $\mathcal{L}_{2D}$ loss based on different keypoints. For instance, the right side lifter incurred a loss due to the discrepancy between the original 2D keypoint coordinates of the right wrist and its 2D coordinates once lifted into 3D, inversely rotated, and re-projected. The $\mathcal{L}_{2D}$  loss for the left side lifter excluded this error, as it does not predict the 3D ordinate for this keypoint. In addition to the 2D loss, a 3D consistency loss is also included to improve the self-consistency within our model. This loss compares the original 3D pose predictions $\hat{\mathbf{Y}}_{3D}$ with the 3D pose obtained when the predictions from the virtual viewpoint are inversely rotated. The 3D consistency loss is given by:
\begin{equation}
\label{threed_loss}
\mathcal{L}_{3D}=\left|\hat{\mathbf{Y}}_{3D}-\mathbf{R}^{-1}\tilde{\mathbf{Y}}_{3D}\right|
\end{equation}
During training, as we were using multiple lifting networks, we produced three 3D poses for each pose of input. The first pose was created by combining the results of the leg and torso network. The second was produced by combining the results of the left and right side lifting networks, with the right side network predictions used for the spine, neck, head and head-top keypoints. The third was produced identically to the second but with the left side lifter predictions used for the spine, neck, head and head-top keypoints. For the first pose the elevation angle used within $\mathbf{R}$ was the average value from the leg and torso lifting networks. For the second and third pose the elevation angle used in $\mathbf{R}$ was the average from the left and right lifting networks.
\subsection{Occlusion Handling Network}
The final stage of our proposed method involved the transfer of knowledge from the independent lifting networks to our occlusion network $O_{3D}$. Specifically, we simulated various occlusion scenarios by masking a 2D pose and then rearranged the lifting networks to obtain a partial 3D pose in this occlusion scenario. Our occlusion network was then trained to predict the 3D coordinates for the occluded part when given this partial 3D pose as input. We trained our occlusion models using knowledge distillation and it learnt to match its own predictions of the missing coordinates with that of the lifting networks if there were no occlusions present. This loss is given by:
\begin{equation}
\begin{split}
\label{occlusion_eq}
\mathcal{L}_{3D occ} = \|(\mathbf{x}_m, \mathbf{y}_m, \hat{\mathbf{z}}_m) - (\hat{\mathbf{x}}_o, \hat{\mathbf{y}}_o, \hat{\mathbf{z}}_o)\|^2 \\
\mathbf{where } \quad (\hat{\mathbf{x}}_o, \hat{\mathbf{y}}_o, \hat{\mathbf{z}}_o) = O_{3D}(\mathbf{x}_p, \mathbf{y}_p, \hat{\mathbf{z}}_p)
\end{split}
\end{equation}
where, $(\mathbf{x}_m, \mathbf{y}_m, \hat{\mathbf{z}}_m)$ represents the 3D coordinate predictions of the missing part by our lifting networks if the occluded part was visible,  $(\hat{\mathbf{x}}_o, \hat{\mathbf{y}}_o, \hat{\mathbf{z}}_o)$ represents the occlusion models predictions of this missing part and $({\mathbf{x}}_p, {\mathbf{y}}_p, {\hat{\mathbf{z}}}_p)$ represents our partial 3D pose given as input to the occlusion model.
\subsection{Additional Losses}
We included two additional losses within our study which have shown to improve results in prior work. \cite{Wandt_2022_CVPR} demonstrated that although many properties of the human body are unknown in an unsupervised setting, relative bone lengths are nearly constant among people \cite{Pietak2013FundamentalRA}. Using this assumption, we calculated the error between the predicted poses' relative bone lengths and a pre-calculated mean (given in \cite{Wandt_2022_CVPR}):
\begin{equation}
\mathcal{L}_{b} = \frac{1}{K}\Sigma_i^K\|b_i - \hat{b}_i\|^2
\end{equation}
where $b_i$ is the pre-calculated relative bone length of bone $i$, $\hat{b}_i$ is our models predicted bone length for bone $i$ and $K$ is the total amount of bones in our 3D pose. Our second loss is that of temporal deformation introduced by \cite{Yu_2021_ICCV}, where they showed that it was beneficial to consider the movement between two poses at different time steps. As we were not dealing with temporal data, we defined the same loss to be the same between two different samples from the training batch:
\begin{equation}
\mathcal{L}_{def} = \|(\hat{\mathbf{Y}}_{3D}^a - \hat{\mathbf{Y}}_{3D}^b) - (\tilde{\mathbf{Y}}_{3D}^a - \tilde{\mathbf{Y}}_{3D}^b)\|^2
\end{equation}
where $\hat{\mathbf{Y}}_{3D}$ and $\tilde{\mathbf{Y}}_{3D}$ are the predicted 3D poses from the real and virtual viewpoint and $a$ and $b$ represent their position in the training batch.
\subsection{Training and Architecture}
As previously stated, we trained four lifting networks that each predicted the 3D coordinates for the leg, torso, and left and right side respectively. The likelihood estimation of each 2D segment obtained from the rotated and reprojected 3D pose comes from a respective pre-trained normalising flow. For the normalising flows we adopted the neural network architecture proposed in \cite{realnvp}, which includes 8 coupling blocks. Each sub-network responsible for predicting the affine transformation was comprised of 2 fully connected layers with 1024 neurons and utilised the ReLU activation function. As for the lifting networks, we drew inspiration from the works of \cite{Martinez_2017_ICCV} and \cite{Wandt_2022_CVPR}. These networks were designed with two paths: one for predicting depth and the other for estimating the elevation angle. Each path consisted of 3 residual blocks with a shared residual block before each path. We trained our lifting networks and flows for 100 epochs with a batch size of 256 using the Adam optimiser with an initial learning rate of $2\times10^{-4}$ which decayed exponentially by 0.95 every epoch. Generative sampling was included within the training of our flows and lifting networks with a $\sigma$ of 0.2. The final objective function for our lifting networks was:
\begin{equation}
\mathcal{L}_{lift} = \mathcal{L}_{NF} + \mathcal{L}_{2D} + \mathcal{L}_{3D} + \mathcal{L}_{def} + 50\mathcal{L}_{b}
\end{equation} 
Our occlusion networks were trained for 10 epochs with the same optimiser and learning rate as the lifting networks. To exploit the rotational coherence within 3D poses and enhance the occlusion handling network's adaptability, we also randomly rotated the partial 3D poses fed into the occlusion handling network. These rotations were executed solely along the azimuth axis and ranged from $-\pi$ to $\pi$.
\section{Results and Evaluation}
\begin{figure}[h]
\centering
{\includegraphics[width=\columnwidth]{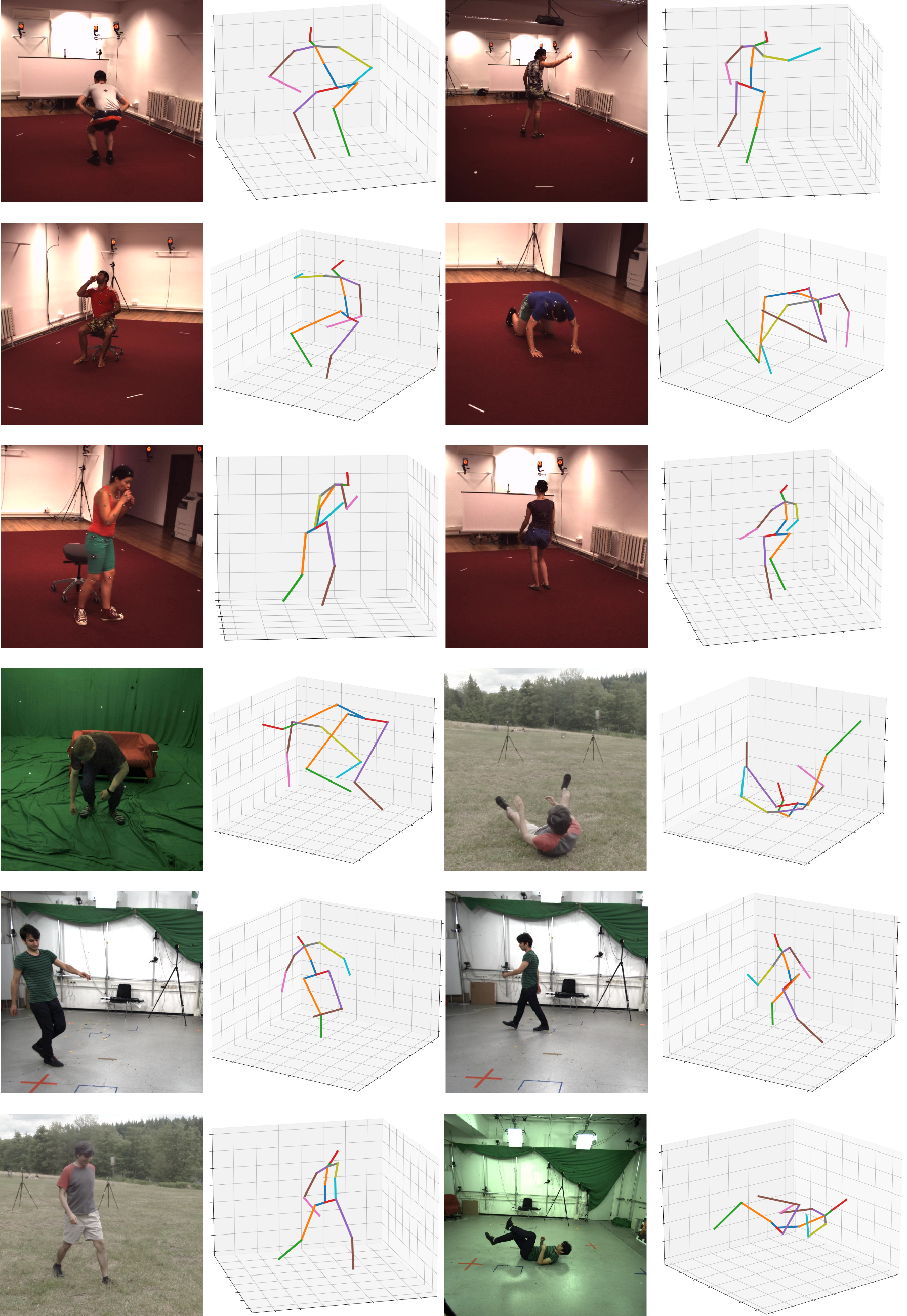}}
\caption{Showing qualitative results obtained from our LInKs model when there is no occlusion present in the scene. The top 3 rows show results from the Human3.6M dataset and the bottom 3 rows show results from the MPI-INF-3DHP dataset.}
\label{qualitative}
\end{figure}

Here we compare the performance of both our lifting networks and occlusion networks on two widely used 3D datasets: Human3.6M \cite{h36m_pami} and MPI-INF-3DHP \cite{mono-3dhp2017}. Our findings indicate that in scenarios without occlusion, our lifting model achieves superior performance on both the Human3.6M and MPI-INF-3DHP datasets for all metrics when compared to prior research. Moreover, we assessed the effectiveness of our approach in handling 2D occlusion on the Human3.6M dataset. Qualitative results of our approach in non-occlusion scenarios can be seen in Figure \ref{qualitative}.
\subsection{Human3.6M Results.} Human3.6M \cite{h36m_pami} is one of the largest and most widely used pose datasets consisting of motion capture data and videos captured from four viewpoints of eleven actors performing diverse actions. The dataset is evaluated using Mean Per Joint Position Error (MPJPE), which is the Euclidean distance in millimetres between the predicted and ground-truth 3D coordinates. Two common protocols are employed when evaluating Human3.6M. The first is N-MPJPE which employs scaling to the 3D predicted pose prior to evaluation. The second is PA-MPJPE where the 3D predicted pose undergoes Procrustes alignment to the GT 3D pose prior to evaluation. Comparing our approach against other approaches with differing levels of supervision (Table \ref{tab:h36m_results}), it can be seen that our approach demonstrates a 7.9\% improvement over the current reference standard in unsupervised pose estimation \cite{Wandt_2022_CVPR} in PA-MPJPE and a 5\% improvement in N-MPJPE. Additionally, our approach outperforms multiple adversarial unsupervised methods \cite{OCR_POSE, amazon_paper_2, Yu_2021_ICCV}.
\begin{table}[h]
\centering
\resizebox{\columnwidth}{!}{%
\begin{tabular}{@{}llcc@{}}
\toprule
Supervision  & Method             & PA-MPJPE $\downarrow$      & N-MPJPE $\downarrow$      \\ \midrule
Full         & Martinez \etal \cite{Martinez_2017_ICCV}           & 37.1          & 45.5          \\
             & Cai \etal \cite{cai}                & 40.2          & 48.8          \\
             & Pavllo \etal \cite{Pavllo2019CVPR_Facebook}   (T)           & 27.2          & 37.2          \\ \midrule
Weak         & Wandt and Rosenhahn \cite{Wandt2019RepNet}             & 38.2          & 50.9          \\
             & Drover \etal \cite{Drover_2018_ECCV_Amazon}             & 38.2          & -             \\
             & Yang \etal \cite{Yang20183DHP}               & 37.7          & 58.6          \\
             & Tung \etal \cite{AIGN}               & 79.0          & -             \\ \midrule
Unsupervised & Chen \etal \cite{amazon_paper_2}               & 58.0          & -             \\
             & Yu \etal \cite{Yu_2021_ICCV} (T)      & 42.0          & 85.3*         \\
             & Wandt \etal \cite{Wandt_2022_CVPR}            & 36.7          & 64.0          \\
             & Wang \etal \cite{OCR_POSE}           & 44.7          & -             \\ \midrule
             & LInKs (Ours)       & \textbf{33.8} & \textbf{61.6} \\ \bottomrule
\end{tabular}%
}
\caption{Evaluation results for the Human3.6M dataset in mm where the input to the model are the GT 2D image keypoints. The bottom section labelled unsupervised shows comparable methods. \cite{OCR_POSE} is the only comparable model that can also handle occlusions. Best results in bold. Numbers are taken from their respective papers. * indicates the use of a scale prior from the dataset. T indicates the use of additional temporal information.}
\label{tab:h36m_results}
\end{table} 
\subsection{MPI-INF-3DHP Results.} MPI-INF-3DHP \cite{mono-3dhp2017} is a markerless MoCap dataset containing the 3D human poses of 8 actors performing 8 different activities. We report the PA-MPJPE result as well as the Percentage of Correct keypoints (N-PCK) and the corresponding area under the curve (AUC). N-PCK is the percentage of predicted coordinates that are within a fixed threshold of 150\emph{mm} to the ground-truth and AUC reports the N-PCK at a range of thresholds between 0-150\emph{mm}. The results of our analyses using this dataset (Table \ref{tab:mpi-results}) once again demonstrated that our approach outperforms previous unsupervised and weakly-supervised approaches that are unable to handle occlusions.
\begin{table}[h]
\centering
\resizebox{\columnwidth}{!}{%
\begin{tabular}{@{}llccc@{}}
\toprule
Supervision  & Method       & PA-MPJPE $\downarrow$      & N-PCK $\uparrow$  & AUC $\uparrow$           \\ \midrule
Weak         & Kundu \etal \cite{kundu2020self}        & 93.9          & 84.6 & 60.8          \\
             & Wandt and Rosenhahn \cite{Wandt2019RepNet}       & -             & 81.8 & 54.8          \\ \midrule
Unsupervised & Chen \etal \cite{amazon_paper_2}     & -             & 71.1 & 36.3          \\
             & Yu \etal \cite{Yu_2021_ICCV}           & -             & 86.2 & 51.7          \\
             & Wandt \etal \cite{Wandt_2022_CVPR}      & 54.0          & 86.0 & 50.1          \\ \midrule
             & LInKs (Ours) & \textbf{49.7} & \textbf{86.3} & \textbf{54.0} \\ \bottomrule
\end{tabular}%
}
\caption{Evaluation results for the MPI-INF-3DHP dataset in scenarios without occlusion. The bottom unsupervised section shows comparable models. }
\label{tab:mpi-results}
\end{table}
\subsection{Results in Occlusion Scenarios.} 
\begin{figure}[h]
\centering
{\includegraphics[width=0.98\columnwidth]{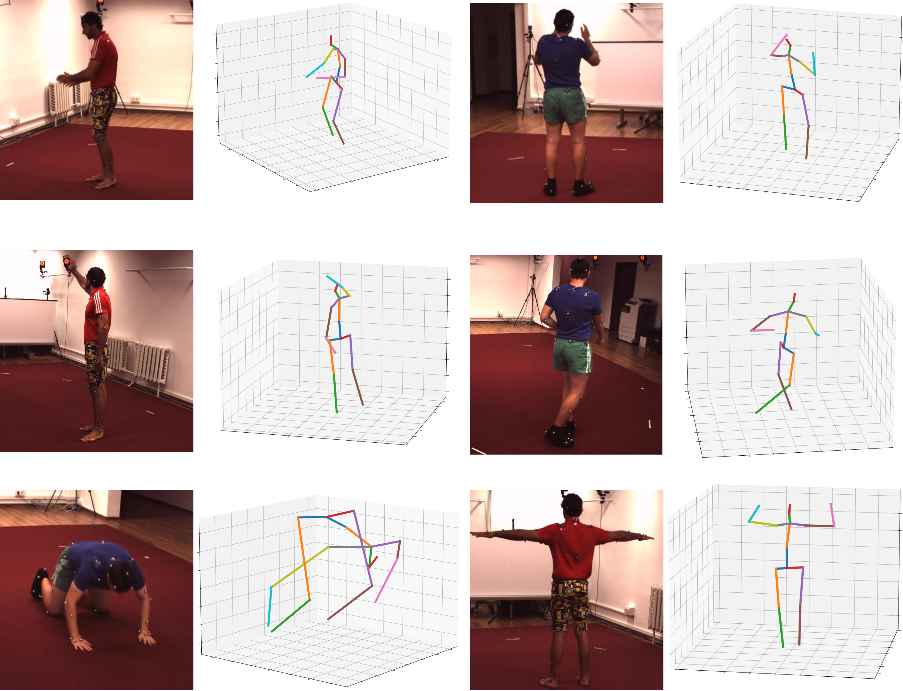}}
\caption{Showing qualitative results from our occlusion handling process on Human3.6M. From top left to bottom right we simulate the occlusion of the right arm, left arm, right leg, left leg, both legs and full torso. The bottom two images show failure cases.}
\label{occlusion_examples}
\end{figure}

To show quantitatively that it is better to first lift an occluded 2D pose to 3D space rather than complete the pose in 2D space, we conducted an additional study on simulated occlusion scenarios on Human3.6M. First, we trained a near-identical occlusion network $O_{2D}$ on the Human3.6M dataset. The only difference between $O_{2D}$ and $O_{3D}$ was that $O_{2D}$ would learn to predict the missing coordinates first in 2D space before the full complete pose was then lifted into 3D. $O_{2D}$ was trained for 10 epochs with the exact same optimiser and learning rate as $O_{3D}$. We trained one network for each type of occlusion evaluated. The results of this experiment can be seen in Table \ref{tab:h36m_results_occlusion} along with the occlusion results of OCR-Pose presented in \cite{OCR_POSE}. Qualitative results of our model in simulated occlusion scenarios can be seen in Figure \ref{occlusion_examples}.
\begin{table}[b]
\centering
\resizebox{\columnwidth}{!}{%
\begin{tabular}{@{}llcc@{}}
\toprule
Model         & Occlusion         & PA-MPJPE $\downarrow$      & N-MPJPE $\downarrow$         \\ \midrule
Wang \etal \cite{OCR_POSE}      & $\mathcal{U}$(0,3) Random Keypoints & 54.8          & -              \\ \midrule
$O_{2D}$ (Ours) & Left Arm          & 61.4          & 85.7           \\
              & Left Leg          & 49.4          & 76.0           \\
              & Right Arm         & 59.8          & 84.5           \\
              & Right Leg         & 51.2          & 75.4           \\
              & Left Arm \& Leg   & 70.7          & 94.9           \\
              & Right Arm \& Leg  & 71.9          & 95.0           \\
              & Both Legs         & 81.5          & 117.0          \\
              & Full Torso        & 96.3          & 136.0          \\ \midrule
$O_{3D}$ (Ours)  & Left Arm          & \textbf{52.1} & \textbf{78.1}  \\
              & Left Leg          & \textbf{46.0} & \textbf{73.2}  \\
              & Right Arm         & \textbf{49.8} & \textbf{75.7}  \\
              & Right Leg         & \textbf{44.5} & \textbf{71.6}  \\
              & Left Arm \& Leg   & \textbf{62.0} & \textbf{86.0}  \\
              & Right Arm \& Leg  & \textbf{60.2} & \textbf{83.7}  \\
              & Both Legs         & \textbf{69.3} & \textbf{99.8}  \\
              & Full Torso        & \textbf{88.4} & \textbf{122.0}
\end{tabular}%
}
\caption{Evaluation results for the Human3.6M dataset in occlusion scenarios. Our 2D MSE model denotes a model with identical parameters trained to fill in the occluded coordinates in 2D space. When a single limb is occluded in our scenarios all keypoints belonging to that limb are occluded e.g. for the arm it will be the shoulder, elbow and wrist.}
\label{tab:h36m_results_occlusion}
\end{table}
As shown, when we compare the two nearly identical occlusion models, one completing the pose in 2D space and the other in 3D space, and the same lifting network, the 3D space occlusion model outperforms the 2D space occlusion model when handling all types of occlusion. It is important to note that the simulated occlusion in OCR-Pose uses random uniform sampling between 0 and 3 when deciding the number of keypoints to occlude, meaning that some poses in their evaluation may actually be unoccluded. In addition, simulating occlusion this way is not realistic as typically specific limbs or segments are missing not just 0-3 random points. Our results show that legs are reconstructed with greater accuracy than arms when using a partial 3D pose, which is consistent with the intuitive understanding that arms possess a higher degree of freedom compared to legs in a 3D context. A surprising find was the fact that our model more accurately located the right arm when it was occluded than the left arm. We hypothesised that the opposite would occur as we assumed people in the Human3.6M dataset were right-handed as the majority of their actions within the dataset were performed with their right arm. Consequently, we expected the right arm to be harder to discern than the left.
\subsection{Ablations} We performed several additional experiments in order to demonstrate the effect of each of our changes on the lifting network. The results of our ablation study in Table \ref{tab:ablation_table} show that each of our changes, including lifting keypoints independently, is important to achieve the best performance. 
\begin{table}[h]
\centering
\resizebox{\columnwidth}{!}{%
\begin{tabular}{@{}lcc@{}}
\toprule
Configuration                     & PA-MPJPE $\downarrow$      & N-MPJPE $\downarrow$  \\ \midrule
 Wandt \etal \cite{Wandt_2022_CVPR}                           & 36.7     & 64.0    \\
Our Recreation of \cite{Wandt_2022_CVPR}       & 36.5     & 65.8    \\
Full Pose Lifter + NF Sampling (Ours)    & 35.1     & 65.2    \\
Left* and Right Lifter + NF Sampling (Ours) & 33.4     & 64.0    \\
Right* and Left Lifter + NF Sampling (Ours) & 33.4     & 64.3    \\ \midrule
Leg and Torso Lifter + NF Sampling (Ours)   & \textbf{33.8}     & \textbf{61.6}    \\ \bottomrule
\end{tabular}%
}
\caption{Ablation study with each of our changes on the Human3.6M dataset. * Indicates which lifting network was used to predict the 3D depth of the spine, neck, head and head-top keypoints.}
\label{tab:ablation_table}
\end{table}Note that we chose to use the predictions of the leg and torso lifter under non-occlusion scenarios as our final results due to their improved performance on N-MPJPE when compared to the left and right lifter.  We hypothesise this is due to the left and right network predictions being at slightly different scales. This occurs in scenarios where one of the subjects' sides is facing the camera making this side appear to have a larger scale in 2D, thus having a larger scale when predicted in 3D. The leg and torso network is able to mitigate this due to seeing both left and right segments of the pose. 
\subsection{Limitations} Our proposed method enables 3D poses to be accurately retrieved even with incomplete 2D information thank to our lift-then-fill approach. The main limitation of this approach however, is that it is not currently capable of accurately dealing with all types of occlusion. For instance, in cases where one keypoint is missing, e.g. the left wrist, our current approach would remove the potentially valuable information of the left shoulder and left elbow 2D coordinates. This is because our partial 3D estimate would be produced using the right side and legs network, both of which do not use these 2D coordinates. Furthermore, if across-body occlusions such as the left wrist and right ankle are present then our approach in its current form would not work. Additionally, we found our leg occlusion networks made some consistent errors such as predicting someone as crouching or lunging during a sitting action. However, we must appreciate the fact that trying to predict the 3D coordinates of the legs from just the 3D of the torso is a highly difficult task. In future work, we plan to address these challenges by looking at a more robust lift-then-fill approach which is able to handle all types of occlusions. 
\section{Conclusion}
In conclusion, our LInKs approach of 2D-3D lifting with partial 3D pose retrieval in occlusion scenarios can reduce the average error on popular 3D human pose datasets. Moreover, our work extends the applications of normalising flows in pose estimation by incorporating generative sampling, which enables the flow to learn a more-defined prior distribution of 2D poses. This, in turn, leads to a stronger likelihood of reconstructed 2D poses during training while also providing additional data to the model. Our approach differs from all prior approaches in that we lift individual parts of the 2D pose independently, specifically for the purpose of occlusion handling where all prior approaches would not work. Furthermore, we show that dealing with occlusion in 2D space is inferior to our occlusion handling process in 3D space. In the future, we plan to address the limitations of our approach and investigate adaptive network structures that can handle inputs with different dimensions to acquire a partial 3D pose in more scenarios. We hope our work can inspire more research investigating the difficult task of occlusion handling in unsupervised 2D-3D lifting.

{\small
\bibliographystyle{ieee_fullname}
\bibliography{egbib}
}

\end{document}